\documentclass[pmlr,twocolumn,10pt]{jmlr} 

\usepackage{booktabs}
\usepackage{siunitx}

\theorembodyfont{\upshape}
\theoremheaderfont{\scshape}
\theorempostheader{:}
\theoremsep{\newline}

\usepackage{bm,amsmath,amssymb}
\usepackage{xspace}
\usepackage{graphicx}
\usepackage{color}
\usepackage{algorithm}
\usepackage{algorithmic}
\usepackage{colortbl}
\usepackage{minitoc}

\usepackage{url}            
\usepackage{booktabs}       
\usepackage{amsfonts}       
\usepackage{nicefrac}       
\usepackage{microtype}      
\usepackage{longtable,lscape}
\usepackage{array,multirow}
\usepackage{wrapfig}
\usepackage{float}

\usepackage[titletoc]{appendix}

\usepackage{enumitem}

\newcommand{\Hb}{\mathbf{H}}

\newcommand{\Kb}{\mathbf{K}}

\newcommand{\Mb}{\mathbf{M}}

\newcommand{\Ob}{\mathbf{O}}

\newcommand{\Qb}{\mathbf{Q}}

\newcommand{\Vb}{\mathbf{V}}
\newcommand{\Wb}{\mathbf{W}}

\newcommand{\sbb}{\mathbf{s}}

\newcommand{\vb}{\mathbf{v}}
\newcommand{\wb}{\mathbf{w}}

\newcommand{\model}{MedGTX+summary\xspace}

\newcommand{\modelunimodal}{Concat\xspace}
\newcommand{\modelgat}{GAT\xspace}
\newcommand{\modelcbert}{ClinicalBERT\xspace}

\newcommand{\modelvts}{MedGTX\xspace}

\newcommand{\modelUniT}{SAnD\xspace}
\newcommand{\modelBiT}{Transformer\xspace}

\newcommand{\summarynetwork}{summary network\xspace}

\jmlryear{2022}
\jmlrworkshop{Conference on Health, Inference, and Learning (CHIL) 2022} 

\title[Graph-Text Multi-Modal Pre-training for Medical Representation Learning]{Graph-Text Multi-Modal Pre-training 
\\
for Medical Representation Learning}

\author{%
 \Name{Sungjin Park}$^1$ \Email{zxznm@kaist.ac.kr}\\
 \Name{Seongsu Bae}$^1$ \Email{seongsu@kaist.ac.kr}\\
 \Name{Jiho Kim}$^1$ \Email{jiho283@kaist.ac.kr}\\
 \Name{Tackeun Kim}$^2$ \Email{tackeun.kim@snu.ac.kr}\\
 \Name{Edward Choi}$^1$ \Email{edwardchoi@kaist.ac.kr}\\
 $^1$\addr KAIST / Daejeon, Republic of Korea \\
 $^2$\addr Department of Neurosurgery, Seoul National University Bundang Hospital / Seongnam, Republic of Korea
}

\begin{document}

\maketitle

\begin{abstract}
As the volume of Electronic Health Records (EHR) sharply grows, there has been emerging interest in learning the representation of EHR for healthcare applications.
Representation learning of EHR requires appropriate modeling of the two dominant modalities in EHR: structured data and unstructured text.
In this paper, we present \modelvts, a pre-trained model for multi-modal representation learning of the structured and textual EHR data.
\modelvts uses a novel graph encoder to exploit the graphical nature of structured EHR data, and a text encoder to handle unstructured text, and a cross-modal encoder to learn a joint representation space.
We pre-train our model through four proxy tasks on MIMIC-III, an open-source EHR data, and evaluate our model on two clinical benchmarks and three novel downstream tasks which tackle real-world problems in EHR data.
The results consistently show the effectiveness of pre-training the model for joint representation of both structured and unstructured information from EHR.
Given the promising performance of \modelvts, we believe this work opens a new door to jointly understanding the two fundamental modalities of EHR data.

\end{abstract}

\paragraph*{Data and Code Availability}
This paper utilizes the MIMIC-III dataset~\citep{johnson2016mimic}, which is available on the PhysioNet repository~\citep{johnson2016physionet}.
Our models and codes are available at the Github repository\footnote{ \url{https://github.com/sjpark9503/kg_txt_multimodal}}.

\section{Introduction}
\label{sec:intro}
\noindent The wide adoption of Electronic Health Records (EHR) by healthcare institutions makes it one of the richest sources of medical data, with huge potential for improving digital medicine when combined with machine learning.
EHR consists of heterogeneous data modalities and two among them account for a large portion of the data: structured data (\textit{e.g.} medication codes) and unstructured textual data (\textit{e.g.} clinical notes).
While the former provides information regarding patient-hospital interaction in a codified and easy-to-process manner, the latter paints a more granular and free-flowing narrative, thus making both inseparable, complementary sources of clinical information~\citep{walsh2004clinician}.

\begin{figure}[t]
    \floatconts{fig:db_overview}
    {\caption{A hospital admission represented in two modalities. 
    \textbf{Graph}: the admission node has multiple children nodes, such as procedure (Px), diagnosis (Dx), and prescription (Rx), each with its literal nodes.
    \textbf{Text}: the discharge summary provides another view on the same admission information.}}
    {\includegraphics[width=\columnwidth]{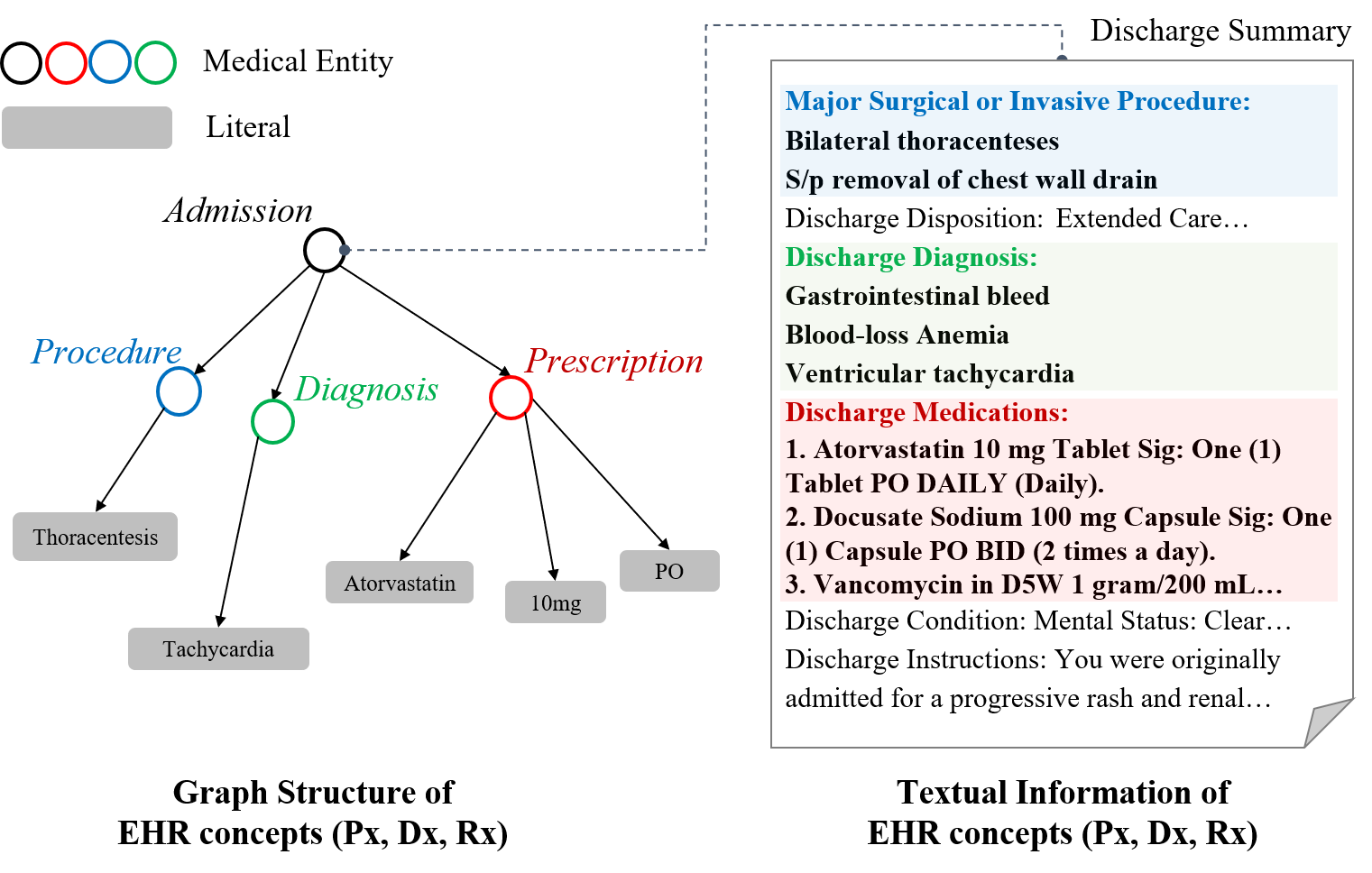}}
\end{figure}
There have been considerable previous works to learn meaningful representations from EHR by either focusing on each modality or straightforwardly merging them.
For structured data understanding, researchers have focused on learning medical concept representations \citep{choi2016learning,choi2017gram,zhang2020hierarchical} and using the underlying graphical structure of medical concepts
\citep{choi2018mime, choi2020learning, hettige2020medgraph}.
In terms of textual data understanding, large scale biomedical pre-trained language models \citep{lee2020biobert, alsentzer2019publicly}
have been the dominant approach due to their state-of-the-art performance on biomedical language understanding tasks.
Some studies used both structured codes and text for various prediction tasks \citep{miotto2016deep, suresh2017clinical, gong2018learning, meng2020hcet}, but they treated both codes and text as bag-of-features which ignore the unique nature of each modality.
Given that textual and structured data in EHR are not just correlated but also complementary to each other, well-designed joint representation learning will provide the opportunity to better capture the heterogeneous nature of EHR. 

In this work, we propose \modelvts, a pre-trained model for joint multi-modal representation learning of the structured and textual EHR data.
Specifically, we interpret structured EHR data as a graph \citep{choi2018mime} based on its hierarchical and heterogeneous nature, therefore propose a graph-text multi-modal learning framework.
Our approach extends the recently developed two-stream multi-modal learning architectures \citep{tan2019lxmert,lu2019vilbert} to the joint representation of graph and text. 
In addition, we develop a \summarynetwork, a novel method to incorporate the textual description of a node into the final graph representation, improving the model performance for all downstream tasks at the cost of small computational overhead.
To maximize the model's multi-modal representation learning for EHR data, we use four different pre-training objectives, namely masked language modeling, masked literal prediction, relation classification, and cross-modal alignment prediction.

Using a publicly available EHR dataset, MIMIC-III~\citep{johnson2016mimic}, we extensively evaluate the proposed method with various baselines on two clinical benchmarks and three novel downstream tasks that require a multi-modal understanding of both graph and text data in EHR: cross-modal retrieval, error detection, and clinical note generation.
In all downstream tasks, \modelvts outperformed all baselines thanks to the combination of pre-training objectives and the proposed \summarynetwork.
To our best knowledge, this is the first work to propose graph-text multi-modal pre-training on EHR with rigorous evaluation on various downstream tasks, demonstrating the necessity of a pre-trained model that jointly considers the two important modalities of EHR data.
Given the consistent and promising performance of \modelvts, we believe this work can open a new door to jointly understanding the two fundamental modalities of EHR data.

\section{Related Works}
\label{sec:relatedworks}
\subsection{Representation Learning for Structured EHR Data}
There have been numerous studies focusing on learning representations of various medical concepts such as diagnosis codes, medication orders,     and hospital visits.

Early studies leveraged the longitudinal nature of EHR~\citep{tran2015learning, choi2016learning, choi2016multi, nguyen2016mathtt} to learn medical code representations from a series of visit records, and further tried to incorporate biomedical domain knowledge~\citep{choi2017gram, ma2018kame, yin2019domain, song2019medical, zhang2020hierarchical}.
More recently, studies have been proposed that try to leverage the underlying structure of EHR, such as focusing on the hierarchical nature between diagnosis and treatment codes~\citep{choi2018mime}, or the graphical and temporal relationship between multiple medical codes~\citep{choi2020learning, hettige2020medgraph}.
Note that there have been studies that used both structured codes and text for various prediction tasks~\citep{miotto2016deep, suresh2017clinical, gong2018learning, meng2020hcet}.
These studies, however, do not fully utilize the unique nature of structured codes or the unstructured text.
Structured codes were treated as bag-of-features without considering their complex graphical structure, and the text was aggregated to topics via topic modeling which made it impossible to align words with codes.
Although prior works demonstrated successful representation learning for their own tasks, there has not been an attempt to jointly learn the representations of graphical and textual EHR data for various multi-modal downstream tasks.

\subsection{Representation Learning for Textual EHR Data}
Representation learning for textual healthcare data has been an emerging area of research in recent years. In early studies, pre-trained word embedding methods~\citep{mikolov2013distributed, pennington2014glove} were widely used to model biomedical texts and applied to various applications, such as named entity recognition \citep{moen2013distributional}, abbreviation disambiguation \citep{liu2015exploiting}, and patient de-identification \citep{dernoncourt2017identification}.

Since the success of pre-trained language models (LMs) such as BERT \citep{devlin2019bert}, several domain-specific pre-trained LMs have also emerged in the clinical domain.
A dominant approach for pre-training is to further train the original BERT over biomedical texts.
Specifically for the healthcare field, some LMs~\citep{alsentzer2019publicly,peng2019transfer} utilize clinical texts for pre-training such as clinical notes from MIMIC-III.
\subsection{Multi-modal Learning via Pre-Training}
Multi-modal representation learning has become one of the promising directions due to richer representation by combining various modalities such as graphs, text, and visual inputs.
In vision-language representation learning, a single stream  \citep{su2019vl, chen2019uniter, li2019visualbert} and two-stream  \citep{lu2019vilbert, tan2019lxmert} Transformer-based architecture with pre-training are proposed to jointly represent the image and text.
Unlike vision-language representations, previous works on graph-text multi-modal learning attempt to enrich one modality by injecting external information from another modality such as text-aware knowledge embedding \citep{xu2017knowledge, xiao2017ssp, xie2016representation} and knowledge-enhanced language models \citep{zhang2019ernie, peters2019knowledge, logan2019barack}.
Although there exist some prior works that focus on jointly aligning the embedding of entity and words \citep{wang2014knowledge, zhong2015aligning, toutanova2015representing}, they need manual alignment between graph and text and were not evaluated on challenging tasks to verify cross-modal understanding.
Recently, similar to our work, GraphCodeBERT \citep{guo2020graphcodebert} proposed a model for learning representations of program source codes, where they jointly embed abstract syntax trees and the comments.

In this work, we focus on the graph-text multi-modal learning for EHR by employing four pre-training tasks and the summary network designed to reflect both the structured and unstructured information in EHR data.
\begin{figure*}[t]
    \floatconts{fig:model_gtxehr}
    {\caption{Overview of the proposed MedGTX model for the joint representation learning of structured and unstructured data in EHR. The model consists of two uni-modal encoders and the cross-modal encoder. We pre-train the model with four objectives: masked language modeling (MLM), masked literal prediction (MLP), relation classification (RC), and cross-modal alignment prediction (AP).}}
    {\includegraphics[width=2.0\columnwidth]{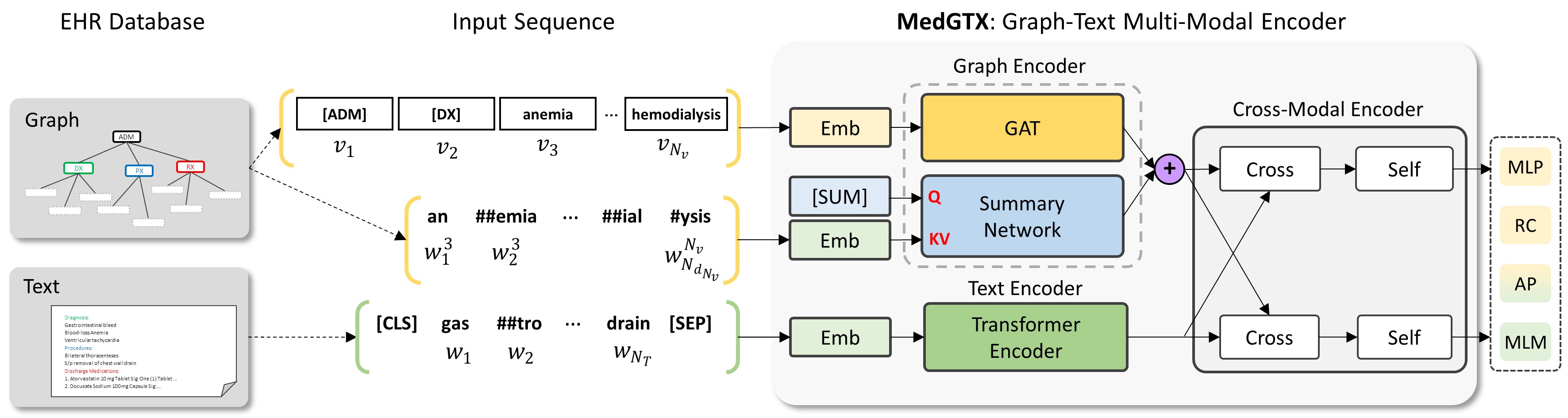}}
\end{figure*}

\section{Methods}
\label{sec:methods}
\subsection{Model Architecture}

\label{ssec:model}
As shown in Figure \ref{fig:model_gtxehr}, our model employs a two-stream architecture where the model separately encodes each input (graph and text) with self-attention layers and combines two modalities in the cross attention layers.
Below, we describe the details of the five main components of \modelvts: the architecture and notations for each modality, attention layer, text encoder, graph encoder, and cross-modal encoder.

\subsubsection{Text Notations}
Following the text preprocessing and embedding strategy of  BERT \citep{devlin2019bert}, we first tokenize the input sentence $W$ into $N_T$ tokens (\textit{i.e.} subwords), $W = \{w_1,w_2, \ldots, w_{N_T}\}$.
We then add special tokens [CLS] and [SEP] at the front and back of $W$ for downstream tasks.
Note that our source of text, the discharge summary, consists of multiple sections, and each section is tagged as respective section headers (\textit{e.g.} "Discharge Medication").
As sections can serve a similar role as the segments in BERT, we build a lookup table of section headers $c \in \{0(\text{Dx}),1(\text{Px}),2(\text{Rx})\}$.
For text embedding, each token $w_i$, its position $i$, and section header $c_i$
are transformed to respective embeddings via the embedding layer, then summed up to produce the final embeddings [CLS], $\wb_1, \ldots, \wb_{N_T}$, [SEP].

\subsubsection{Graph Notations}
Given a EHR graph $G=(V,E)$, $V=\{v_1, \dots, v_{N_v}\}$ is a set of nodes with $N_v$ denoting the number of nodes, and $D = \{d_1, ..., d_{N_v}\}$ is a set of textual descriptions in a graph where $d_i$ belongs to $v_i$, and the description consists of $N_{d_i}$ tokens $d_i = w^i_1, ...,w^i_{N_{d_i}}$ and $E = \{e_1, ..., e_{N_e}\}$ is a set of relations with $N_e$ denoting the number of relations in the graph $G$.
Our EHR graph is converted from the relational database, therefore some nodes (\textit{i.e. abstract nodes}) represent abstract concepts such as a single admission or a single medication order, while other nodes (\textit{i.e. literal nodes}) represent concrete values such as
a medication volume (``250mg'').
Based on this observation, we use special tokens ([ADM], [DX], [PX], [RX]) to represent the abstract nodes while literal nodes are represented with unique embeddings. 
Furthermore, abstract nodes' $d_i$ (\textit{e.g.} admission ID) are unused throughout the paper as they provide little information to the model.
The difference between abstract nodes and literal nodes are further discussed in Appendix~\ref{supp:absLiteralNodes}.

For graph embedding, each node $v_i$ and its position $i$ are transformed to respective embeddings via the embedding layer, then summed up to produce the final embeddings $\vb_1, \ldots, \vb_{N_v}$.
Note that position embeddings are required for the model to distinguish multiple abstract nodes of the same type (\textit{e.g.} a single admission contains multiple medication orders).
\subsubsection{Multi-Head Attention Layer}
We define the multi-head attention mechanism~\citep{vaswani2017attention} which will be used in all uni-modal and cross-modal encoders.
We use the canonical scaled dot-product as follows, where we omit the multi-head component to reduce clutter, 
\begin{align*}
    \Qb = \Hb_Q \Wb_Q, \enskip \Kb = \Hb_{KV} \Wb_K, \enskip \Vb = \Hb_{KV} \Wb_V \\
    \text{Attn}\Big(\Hb_Q, \Hb_{KV}, \Mb \Big) = \text{softmax}\bigg(\frac{\Qb \Kb^\top}{\sqrt{d_k}}+\Mb\bigg) \Vb
\end{align*}
where $\Hb_Q$ denotes the input features for queries, $\Hb_{KV}$ the input features for keys and values, $d_k$ the dimension size of the key vectors, $d_v$ the dimension size of the value vectors, and $d_{h}$ the dimension size of hidden states.
$\Wb_Q, \Wb_K \in \mathbb{R}^{d_{h} \times d_k}$, and $\Wb_V \in \mathbb{R}^{d_{h} \times d_v}$ are trainable projection weights for query, key, and value.
$\Mb$ denotes an attention mask comprised of 0s and negative number $n$s for indicating which features are allowed to attend to one another.

\subsubsection{Text Encoder}
The text encoder consists of $L_T$ layers of Transformer encoders, where the first layer takes text embeddings $\Hb^{(0)}_{T} = \mbox{[CLS]}, \wb_1, \ldots, \wb_N, \mbox{[SEP]}$.
Transformer encoders are then repeatedly applied, which consist of the multi-head self-attention and the position-wise feed-forward network.
The output $\Hb^{(l)}_{T}$ of the $l$-th text encoder is obtained as follows,
\[\Hb^{(l)}_{T} = \text{Attn}\Big(\Hb^{(l-1)}_{T}, \Hb^{(l-1)}_{T}, \Mb_{pad}\Big)\]
where $\Mb_{pad}$ consists of $n$s at padding token positions to exclude it from attention score, and 0s elsewhere.

\label{subsubsec:graphEnc}
\subsubsection{Graph Encoder}
The graph encoder consists of $L_G$ layers of two separate components: the graph attention network (GAT)~\citep{velivckovic2017graph} and the \summarynetwork.
GAT encodes the graphical structure of the EHR graph by considering the adjacency between all pairs of nodes. Starting with $\Hb^{(0)}_{G} = \vb_1, \ldots, \vb_{N_v}$, at $l$-th GAT layer, the previous output $\Hb^{(l-1)}_{G}$ is transformed as follows,
\[\Hb^{(l)}_{G} = \text{Attn}\Big(\Hb^{(l-1)}_{G}, \Hb^{(l-1)}_{G}, \Mb_{adj}\Big)\]
where $\Mb_{adj}$ is a symmetric adjacency matrix that reflects the EHR graph structure.

A common approach to introduce node features $D$ into GAT is initializing each node embedding $\vb_i$ with the mean of token embeddings of $d_i$. 
However, EHR graphs often contain repetitive and redundant words (\textit{e.g.} ``main'', ``IV'') in $d_i$, and the importance of each word $w^{i}_{j}$ depends on the downstream task.
To this end, we propose a novel summary network that extracts keywords from $D$ and encodes them into a fixed-size summary vector $\sbb$ independent of GAT.
We add a special token [SUM] to the node lookup table and its embedding vector is used as the input $\sbb^{(0)}$ to the summary network.
At $l$-th layer of the summary network, the previous output $\sbb^{(l-1)}$ is transformed as follows,
\[\sbb^{(l)} = \text{Attn}\Big(\sbb^{(l-1)}, \Hb_{D}, \Mb_{pad}\Big)\]
where $\Hb_D=\wb^1_1,\wb^1_2,\ldots,\wb^{N_v}_{N_{d_{\small{N_v}}}}$ is a concatenation of all descriptions in the graph.
Note that the purpose of the \summarynetwork is the extraction of keywords rather than contextualization of the unstructured node features in the graph, so we directly inject the low-level text features as a key and value of the \summarynetwork.
We empirically demonstrate the superiority of our summary network over the node initialization technique in Section \ref{ssec:abaltion}. 

\subsubsection{Cross-Modal Encoder}
We apply the cross-modal encoder on the text features $\Hb^{(L_T)}_{T}$ and the graph features $\Hb^{(L_G)}_{G}$ and $\sbb^{(L_G)}$.
The cross-modal encoder~\citep{tan2019lxmert} consists of $L_C$ cross-attention layers, and $L_S$ self-attention layers, and position-wise feed-forward networks.
The cross-attention is equivalent to the encoder-decoder attention~\citep{vaswani2017attention}, in that the key and value vectors come from the other modality (Fig.~\ref{fig:model_gtxehr}).
The output of the $l$-th cross-attention layer is,

\begin{align*}
    \Ob^{(l)}_{T} = \text{Attn}\Big(\Ob^{(l-1)}_{T}, \Ob^{(l-1)}_{G}, \Mb_{pad}\Big) \\
    \Ob^{(l)}_{G} = \text{Attn}\Big(\Ob^{(l-1)}_{G}, \Ob^{(l-1)}_{T}, \Mb_{pad}\Big) 
\end{align*}
where $\Ob^{(0)}_{T}=\Hb^{(L_T)}_{T}$, $\Ob^{(0)}_{G}=[\sbb^{(L_G)}; \Hb^{(L_G)}_{G}]$, and $\Ob^{(l)}_{T}$ is a text output which is blended with graph features, $\Ob^{(l)}_{G}$ vice versa.
Note that $[\sbb^{(L_G)}; \Hb^{(L_G)}_{G}]$ is a concatenation of $\sbb^{(L_G)}$ and $\Hb^{(L_G)}_{G}$ along the sequence.
Here we assume that the summary network was used. We use $\sbb^{(0)}$ instead of $\sbb^{(L_G)}$ when we omit the summary network, which is similar to [CLS] in the text modality.
The process for the self-attention layers and the feed-forward networks are the same as uni-modal encoders.

\begin{table}[t]
\floatconts
    {tab:data_stat}
    {\caption{Data statistics of Dx+Px and Rx. Train/Valid/Test split for \# samples. \# nodes is the sum of abstract nodes and literal nodes.}}
    {
        \resizebox{\columnwidth}{!}{%
            \begin{tabular}{ccccc}
            \toprule
            Dataset
            & \# samples
            & \# nodes
            & \# relations
            & \# literals\\
            \midrule
            \multirow[m]{1}{*}{Dx+Px}
            & 28,915\;/\;2,000\;/\;2,000 & 699,724 & 6 & 8,049\\
            \multirow[m]{1}{*}{Rx}
            & 23,663\;/\;1,000\;/\;1,000 & 3,015,345 & 7 & 9,291\\
            \bottomrule
            \end{tabular}
        }
    }
\end{table}

\subsection{Pre-training Objectives}
\label{ssec:pretrain}
We describe four pre-training objectives used for training \modelvts.
Further pre-training details can be found in Appendix~\ref{supp:PreSetting}.
\paragraph{Masked Language Modeling (MLM)}
We use the masked language modeling (MLM) task from BERT to pre-train for the text modality.
A random sample of the tokens in the text input sequence is selected and replaced with the special token [MASK]. The MLM objective is a cross-entropy loss on predicting the masked tokens.
Unlike BERT, \modelvts is encouraged to leverage both textual context and graph information to recover the masked tokens correctly.
During pre-training, words in the input sequence are randomly masked with a probability of 0.15. We replace the chosen token with (1) the [MASK] token with 80\% chance (2) a random token with 10\% chance (3) the original token with 10\% chance.
\paragraph{Masked Literal Prediction (MLP)}
Similar to MLM, we randomly mask literal nodes of the EHR graph with a probability of 0.15 and pre-train the model to recover its identity as another way to encourage \modelvts to leverage both structured and unstructured modalities.
Unlike MLM, we do not replace the nodes with a random node since the graph encoder considers the node connections.
\paragraph{Relation Classification (RC)}
As there are multiple relation types in the EHR graph (\textit{e.g. MedicationOrder, Diagnosis, DiagnosisName, DrugRoute}), which were not explicitly injected into the model architecture, we propose relation classification to learn the multi-relational aspect of structured EHR.
Specifically, we randomly sample 10\% of node pairs in the graph, concatenate their corresponding output contextualized features, and predict their relation type, including \textit{Not Connected}.
\paragraph{Cross-modal Alignment Prediction (AP)}
We follow the cross-modal alignment prediction task in LXMERT \citep{tan2019lxmert}.
The purpose of the cross-modal alignment prediction objective is to encourage the model to align a representation space of the graph and language modality.
First, for each graph-text pair, we replace it with a text from another admission with the probability of 0.5.
Then, we concatenate the special tokens [SUM] and [CLS] to classify whether the given graph-text pair comes from the same admission or not.

\begin{table*}[t]
\floatconts
    {tab:retrieval_result}
    {\caption{Test-set results on text \& graph retrieval. We report Hits@10 and mean reciprocal rank.}}
    {\resizebox{2.0\columnwidth}{!}{
        {\small
        \begin{tabular}{lcccccccc}
        \toprule
        \multicolumn{1}{c}{\multirow{3}[1]{*}{Model}} & \multicolumn{4}{c}{Text Retrieval} & \multicolumn{4}{c}{Graph Retrieval}\\
        \cmidrule(lr){2-5}
        \cmidrule(lr){6-9}
        &\multicolumn{2}{c}{Dx+Px} & \multicolumn{2}{c}{Rx} & \multicolumn{2}{c}{Dx+Px} &\multicolumn{2}{c}{Rx}\\
        \cmidrule(lr){2-3}
        \cmidrule(lr){4-5}
        \cmidrule(lr){6-7}
        \cmidrule(lr){8-9}
        & Hits@10 & MRR & Hits@10 & MRR & Hits@10 & MRR & Hits@10 & MRR\\
        \midrule
        \modelunimodal
        & 51.6 (0.3)& 24.1 (0.4) & 74.0 (2.9)& 43.3 (4.2) & 52.0 (0.8)& 24.8 (0.5) & 74.8 (0.7)&45.5 (1.5)\\
        \modelvts
        & 76.1 (0.6) & 47.2 (0.3) & 93.0 (0.2) & 75.8 (1.4)
        & 76.4 (0.5) & 47.8 (0.4) & 92.1 (0.6) & 77.4 (1.1)\\
        \hspace{3mm}-\;\;PT
        & \enskip2.0 (0.4) & \enskip1.3 (0.2) & \enskip1.0 (0.0) & \enskip0.8 (0.0)
        & \enskip2.0 (0.7) & \enskip1.3 (0.3) & \enskip1.1 (0.1) & \enskip0.8 (0.1)\\
        \hspace{3mm}+\;summary
        & \textbf{77.4 (0.3)} & \textbf{51.4 (0.6)} & \textbf{96.2 (0.1)} & \textbf{87.7 (0.1)}
        & \textbf{78.2 (0.4)} & \textbf{52.8 (0.4)} & \textbf{95.5 (0.1)} & \textbf{86.9 (0.1)} \\
        \bottomrule
      \end{tabular}%
      }}
    }
\end{table*}
\begin{table}[t]
    \floatconts
    {tab:mortality_result}
    {\caption{Test-set results on mortality prediction. We report Area Under Precision-Recall Curve.}}
    {\resizebox{\columnwidth}{!}{
        {\small
        \begin{tabular}{clcc}
        \toprule
        Modality & \multicolumn{1}{c}{Model} & Dx+Px & Rx\\
        \midrule
        \multirow[m]{4}{*}{Uni-Modal}
        & \modelcbert
        & 12.3 (1.8) & 37.8 (3.4)\\
        & \modelgat
        & 20.0 (2.3) & 26.7 (4.1) \\
        & \modelBiT
        & 19.6 (1.4) & 25.8 (3.4) \\
        & \modelUniT
        & - & 10.0 (2.1) \\
        \midrule
        \multirow[m]{4}{*}{Multi-Modal}
        & \modelunimodal
        & 19.5 (1.1) & 41.4 (1.4)\\
        & \modelvts
        & 21.2 (0.5)& 43.2 (2.5)\\
        & \hspace{3mm}-\;\;PT
        & 19.2 (1.4) & 37.1 (0.7)\\
        & \hspace{3mm}+\;summary
        & \textbf{24.1 (0.8)} & \textbf{43.9 (1.5)}\\
        \bottomrule
      \end{tabular}}
      }
    }
\end{table}

\section{Experiments}
\label{sec:experiments}
In this section, we first introduce the dataset used for pre-training and downstream evaluation, followed by the baseline model description.
We then describe five downstream tasks: two existing clinical benchmarks and three novel multi-modal downstream tasks, along with experimental results.
A detailed description of experimental settings for each downstream task can be found in Appendix~\ref{supp:exp_setup}.

\subsection{Datasets}
\label{sec:datasets}
We evaluate all models on MIMIC-III~\citep{johnson2016mimic}, a freely-available multi-modal EHR dataset.
Collected for approximately 40,000 ICU patients at Beth Israel Deaconess Medical Center between 2001 and 2012, MIMIC-III consists of heterogeneous data such as diverse structured information (\textit{e.g.} diagnoses, prescriptions, procedures) and textual data (\textit{e.g.} physician notes, nursing notes, discharge summaries).
We convert MIMIC-III to a graph from a relational database by following the approach
in MIMICSQL* ~\citep{park2020knowledge}. We describe the details of our EHR graph in Appendix~\ref{supp:data_proc}.
Specifically for text data, we use discharge summaries, which contain a summary of the medical events that occurred during single hospital admission (Figure \ref{fig:db_overview}).
Given that discharge summaries are divided into sections
, we can extract text regarding the diagnoses (Dx), procedures (Px), and prescriptions (Rx) that occurred during the admission.
As an initial step towards apply our framework to the entire EHR, we merge datasets which share a certain level of similarity. 
Specifically, Dx and Px both share the exactly same table schema in MIMIC-III.
Furthermore, the number of records in Dx and Px are comparatively smaller than Rx. Therefore, we merge Dx and Px into a single dataset Dx+Px without changing the size and characteristics. 
We also considered merging Dx, Px, and Rx into a single dataset, Dx+Px+Rx, but this would significantly increase the text length and the number of nodes in a graph. 
Due to the limitation of computational resource, we keep the prescriptions as a separate dataset Rx, and leave the development of an efficient approach to encode a large graph and text from EHR as a future work.
A summary statistics of the two datasets is provided in Table \ref{tab:data_stat}. 

\subsection{Models}
\begin{itemize} [leftmargin=3mm]
    \item \textbf{Uni-modal baselines}
    \enskip To verify the effectiveness of the multi-modal joint representation, we evaluate uni-modal baselines on downstream tasks.
    For the text modality, we employ a biomedical pre-trained language model, ClinicalBERT~\citep{alsentzer2019publicly}, which is trained on PubMed and MIMIC-III.
    For the graph modality, we employ GAT~\citep{velivckovic2017graph} (or Transformer encoder) to encode the graph with (or without) the structural information, respectively. 
    Furthermore, we also compare with SAnD~\citep{song2018attend} which interprets our graph as a time series.
    We evaluate SAnD only on Rx because Dx+Px does not have temporal information.
    For the note generation task, we choose a simple sequence-to-sequence (Seq2Seq) as the uni-modal baseline, which encodes the graph with a Transformer encoder and decodes the text with a Transformer decoder.
    \item \textbf{\modelunimodal}
    \enskip We replace cross attention layers in \modelvts with self-attention layers, equivalent to two independent single modality encoders. For a fair comparison, \modelunimodal is pre-trained with the same objectives as \modelvts. This baseline is used to verify the effect of the joint representation learning of graph and text modalities.
    \item \textbf{\modelvts} \enskip Both \modelvts and MedGTX+summ
    \newline ary are our pre-trained models, but the graph encoder of \modelvts does not contain the summary network proposed in Section~\ref{subsubsec:graphEnc}.
    \textit{-\;PT} is proposed to verify the effectiveness of the pre-training, where we randomly initialize \modelvts and directly fine-tune it for downstream tasks.
\end{itemize}
In all downstream tasks, all models were trained with five random seeds, and we report the mean and the standard deviation.
We use all four pre-training tasks for the Dx+Px dataset, while we exclude the relation classification (RC) task for the Rx dataset as all models showed better performance without it.
Also, we initialize the text encoder with parameters from BERT-tiny~\citep{turc2019wellread}. 

\subsection{Cross-Modal Retrieval}
Cross-modal retrieval aims to find the most relevant one among a pool of text given a graph (\textit{i.e.} Text Retrieval) or vice versa (\textit{i.e.} Graph Retrieval).
This task serves as a sanity check for the cross-modal alignment objective.
At inference, given pairs are ranked based on their respective prediction scores.
We use Hits@10 and mean reciprocal rank (MRR) to report the model performance for both text retrieval and graph retrieval. Hits@k describes the fraction of true graph (or text) that appear in the first k graphs (or text) of the sorted rank list, and MRR is the inverse of the harmonic mean of the ranks.

Table~\ref{tab:retrieval_result} shows the model performance on the Dx+Px and Rx datasets. 
Results show that for all models, the proposed pre-training objectives dramatically increase the retrieval performance for all cases.
The summary network also leads to a dramatic increase of MRR in both retrieval tasks compared to \modelvts, indicating the effectiveness of the proposed model architecture.
\begin{table}[t]
  \floatconts
  {tab:readm_result}
  {\caption{Test-set results on readmission prediction.
  We report Area Under Precision-Recall Cur\\ve.}}
  {\resizebox{\columnwidth}{!}{%
      {\small
      \begin{tabular}{clcc}
        \toprule
        Modality & \multicolumn{1}{c}{Model} & Dx+Px & Rx\\
        \midrule
        \multirow[m]{4}{*}{Uni-Modal}
        & \modelcbert & 32.2 (1.6)  & 34.8 (3.0)\\
        & \modelgat   & 36.5 (1.8)  & 37.4 (2.3) \\
        & \modelBiT   & 35.8 (1.3)  & 35.0 (3.2) \\
        & \modelUniT  & -           & 28.6 (1.1) \\
        \midrule
        \multirow[m]{4}{*}{Multi-Modal}
        & \modelunimodal            & 36.1 (1.1) & 41.0 (2.3)\\
        & \modelvts                 & 38.4 (0.7) & 38.0 (1.5) \\
        & \hspace{3mm}-\;\;PT       & 37.0 (1.3) & 34.2 (2.3) \\
        & \hspace{3mm}+\;summary    & \textbf{39.6 (0.8)} & \textbf{41.9 (1.4)}\\
        \bottomrule
      \end{tabular}}
      }
    }
\end{table}

\subsection{Temporal Prediction} 
Predicting mortality and hospital readmission are two of the most common clinical application tasks, as they may improve efficiency and reduce the burden of clinical experts.
The prediction target of mortality prediction is either dead (1) or alive (0), and either readmitted (1) or not (0) for readmission prediction.
The prevalence rate of mortality and readmission is 4\% and 27\%, respectively.
We construct two benchmark datasets based on MIMIC-III and report the area under the precision-recall curve (AUPRC) as a performance metric.
Note that MedGTX is trained with discharge summaries, so if we predict in-hospital mortality, that would be leaking information from the future. Therefore, instead, we set up the mortality prediction task to predict whether the patient would die within 30 days after discharge (similar to predicting the 30-day readmission). 
\begin{table}[t]
  \floatconts
  {tab:errdet_result}
  {\caption{Test-set results on error detection. We report F1 score.}}
  {\resizebox{\columnwidth}{!}{%
      {\scriptsize
      \begin{tabular}{clcc}
        \toprule
        Modality & \multicolumn{1}{c}{Model} & Dx+Px & Rx\\
        \midrule
        \multirow[m]{1}{*}{Uni-Modal}
        & \modelgat
        & 80.0 (0.2) & 89.6 (0.1)\\
        \midrule
        \multirow[m]{4}{*}{Multi-Modal}
        & \modelunimodal 
        & 84.1 (0.3) & 99.1 (0.0)\\
        & \modelvts
        & 84.3 (0.3) & 99.1 (0.1)\\
        & \hspace{3mm}-\;\;PT
        & 83.8 (0.3) & 95.9  (0.1) \\
        & \hspace{3mm}+\;summary
        & \textbf{84.6 (0.2)} & \textbf{99.1 (0.1)}\\
        \bottomrule
      \end{tabular}}
      }
  }
\end{table}      

Table~\ref{tab:mortality_result} and~\ref{tab:readm_result} show the performance of mortality and readmission prediction, respectively.
We can observe the trend that using both structured and unstructured modalities generally improves AUPRC compared to uni-modal baselines, and that our pre-training objectives indeed are helpful for clinical predictions.
Furthermore, the description summary of a graph in \modelvts has an advantage even when the task requires prediction over time dimension. 

\subsection{Error Detection}
EHR often contains misspelled values which can lead to critical situations, therefore detecting such errors in advance is an important application.
As structured values are often written based on clinical notes, a chance of error is higher for the graph modality than the text.
Therefore, given a graph-text pair, we replace some literal nodes with another random literal node and train all models to predict if any node was altered.
Since finding as many errors as possible with high precision is the objective of this task, we use F1 score to report the model performance. 
Table~\ref{tab:errdet_result} shows the error detection performance on the Dx+Px and Rx datasets.
Again, it can be seen that the multi-modal signal improves performance for all models in both datasets, showing error detection also makes use of clues from another modality. However, given the similar performance of \modelunimodal and \modelvts, we can infer that a simple multi-modal pre-training task is enough to produce good performance even without explicit cross-attention. 

We discover that most model performances are saturated to a similar point.
We believe this is due to the random replacement of nodes in a graph yielding some trivial errors, and leave the generation of non-trivial errors as the future work.

\subsection{Clinical Note Generation}
\label{ssec:generation}
Clinical note generation aims to translate EHR graph into clinical free text, similar to image captioning in the vision-language domain.
Compared to other downstream tasks, this one evaluates a different aspect of multi-modal learning, namely the generation capability.
Since generation tasks require a decoder, we adopt a UniLM style~\citep{NEURIPS2019_c20bb2d9} fine-tuning and decoding strategy to train all models to generate text without changing the overall model architecture.
We adopt ROUGE (RG) score as the evaluation metric where RG-2 measures the overlap of bigrams and RG-L measures the longest common subsequence (LCS) between the system output and reference.
As shown in Table~\ref{tab:notegen_result}, both datasets enjoy significantly improved performance thanks to our pre-training objectives, and further smaller performance boost thanks to the summary network.
\begin{table}[t]
\floatconts
    {tab:notegen_result}
    {\caption{Test-set results on clinical note generation. We report ROUGE-2/L F1 score.}}
    {\resizebox{\columnwidth}{!}{%
  \begin{tabular}{clcccc}
    \toprule
    \multirow{2}{*}{Modality} & \multicolumn{1}{c}{\multirow{2}{*}{Model}} & \multicolumn{2}{c}{Dx+Px} & \multicolumn{2}{c}{Rx}\\
    \cmidrule(lr){3-4}
    \cmidrule(lr){5-6}
    & & RG-2 & RG-L & RG-2 & RG-L \\
    \midrule
    \multirow[m]{1}{*}{Uni-Modal}
    & Seq2Seq
    & \enskip 6.8 (0.1) & 20.2 (0.1)
    & 33.0 (0.4) & 52.8 (0.3) \\
    \midrule
    \multirow[m]{3}{*}{Multi-Modal}
    & \modelvts
    & \enskip 7.4 (0.1) & 20.6 (0.1)
    & 43.8 (0.2) & 54.1 (0.2)\\ 
    & \hspace{3mm}-\;\;PT
    & \enskip 3.3 (0.2) & 13.9 (0.5) 
    & 34.4 (0.4) & 44.5 (0.7) \\
    & \hspace{3mm}+\;summary
    & \textbf{\enskip 7.5 (0.1)} & \textbf{20.8 (0.2)}
    & \textbf{44.1 (0.2)} & \textbf{54.7 (0.2)} \\
    \bottomrule
  \end{tabular}
  }}
\end{table}

\section{Analysis}
\label{sec:analysis}
In this section, we analyze \modelvts from the perspective of some alternative design choices.
We then perform ablation studies over model components and pre-training objectives.
We also qualitatively assess cross-modal attention and analyze its clinical meaning.
We visualize additional qualitative samples in Appendix~\ref{supp:att_viz}.

\begin{table*}[t]
    \centering
    \caption{Ablation over design choices. 
    We report AUPRC for temporal predictions, F1 for error detection, ROUGE-L for generation, and mean reciprocal rank (MRR) for text retrieval.
    }
    \resizebox{2.0\columnwidth}{!}{
    \begin{tabular}{lcccccccccc}
    \toprule
    \multicolumn{1}{c}{\multirow{2}[1]{*}{Model}}
    & \multicolumn{2}{c}{Mortality}
    & \multicolumn{2}{c}{Readmission}
    & \multicolumn{2}{c}{Error Detection}
    & \multicolumn{2}{c}{Generation}
    & \multicolumn{2}{c}{Text Retrieval}\\
    \cmidrule(lr){2-3}
    \cmidrule(lr){4-5}
    \cmidrule(lr){6-7}
    \cmidrule(lr){8-9}
    \cmidrule(lr){10-11}
    & Dx+Px & Rx
    & Dx+Px & Rx
    & Dx+Px & Rx
    & Dx+Px & Rx
    & Dx+Px & Rx\\
    \midrule
    MedGTX
    & 21.2 (0.5) & 43.2 (2.5)
    & 38.4 (0.7) & 38.0 (1.5)     
    & 84.3 (0.3) & 99.1 (0.1)    
    & 20.6 (0.1) & 54.1 (0.2)     
    & 47.2 (0.3) & 75.8 (1.4)
    \\
    \hspace{3mm}+\;summary
    & \textbf{24.1 (0.8)} & \textbf{43.9 (1.5)}     
    & \textbf{39.6 (0.8)} & \textbf{41.9 (1.4)}     
    & \textbf{84.6 (0.2)} & \textbf{99.1 (0.1)} 
    & 20.8 (0.2) & 54.7 (0.2) 
    & 51.4 (0.6) & 87.7 (0.1)
    \\
    \hspace{3mm}+\;init
    & 22.4 (0.8) & 39.9 (1.4)     
    & 37.2 (1.6) & 38.5 (2.3)     
    & 81.9 (0.4) & 96.6 (0.1)    
    & \textbf{21.2 (0.1)} & 54.3 (0.2)     
    & \textbf{59.7 (0.4)} & \textbf{89.2 (0.3)}
    \\
    \hspace{3mm}-\;\;structure
    & 21.8 (0.6) & 41.5 (1.1)     
    & 36.2 (1.9) & 40.0 (1.8)     
    & - & - 
    & 20.1(0.2) & \textbf{56.3 (0.4)}      
    & 41.2 (2.0) & 86.7 (0.1)
    \\
    \bottomrule
    \end{tabular}
    }

    \label{tab:ablation_model}
\end{table*} 
\begin{table*}[t]
    \centering
    \caption{Ablation over model components.
    We report AUPRC for temporal predictions, F1 for error detection, ROUGE-L for generation, and mean reciprocal rank (MRR) for text retrieval.}
    \resizebox{2.0\columnwidth}{!}{%
    \begin{tabular}{ccccccccccc}
    \toprule
    \multirow{2}[1]{*}{Model}
    & \multicolumn{2}{c}{Mortality}
    & \multicolumn{2}{c}{Readmission}
    & \multicolumn{1}{c}{Error Detection}
    & \multicolumn{2}{c}{Generation}
    & \multicolumn{2}{c}{Text Retrieval}
    \\
    \cmidrule(lr){2-3}
    \cmidrule(lr){4-5}
    \cmidrule(lr){7-8}
    \cmidrule(lr){9-10}
    & Dx+Px & Rx
    & Dx+Px & Rx
    & Dx+Px 
    & Dx+Px & Rx
    & Dx+Px & Rx\\
    \midrule
    \model
    & \textbf{24.1 (0.8)} & 43.9 (1.5)  
    & \textbf{39.6 (0.8)} & \textbf{41.9 (1.4)} 
    & \textbf{84.6 (0.2)} 
    & \textbf{20.8 (0.2)} & 54.7 (0.2) 
    & \textbf{51.4 (0.6)} & \textbf{87.7 (0.1)} 
    \\
    - GAT
    & 22.5 (2.3) & \textbf{44.1 (1.5)} 
    & 36.4 (1.5) & 37.8 (1.5) 
    & 84.3 (0.3) 
    & 20.4 (0.2) & \textbf{54.9 (0.2)} 
    & 36.6 (0.1) & 80.9 (0.3) 
    \\
    - LM init
    & 22.4 (0.9) & 40.0 (1.7) 
    & 36.2 (1.2) & 40.9 (1.2) 
    & 84.5 (0.2) 
    & 20.1 (0.1) & 50.7 (0.2) 
    & 35.5 (1.0) & 62.0 (0.5) 
    \\
    \bottomrule
    \end{tabular}
    }

    \label{tab:ablation_components}
\end{table*} 
\begin{table*}[t]
    \centering
    \caption{Ablation over pre-training tasks. 
    We report AUPRC for temporal predictions, F1 for error detection, ROUGE-L for generation, and mean reciprocal rank (MRR) for text retrieval.
    }
    \resizebox{2.0\columnwidth}{!}{%
    \begin{tabular}{ccccccccccc}
    \toprule
    \multirow{2}[1]{*}{Pre-training task}
    & \multicolumn{2}{c}{Mortality}
    & \multicolumn{2}{c}{Readmission}
    & \multicolumn{1}{c}{Error Detection}
    & \multicolumn{2}{c}{Generation}
    & \multicolumn{2}{c}{Text Retrieval}
    \\
    \cmidrule(lr){2-3}
    \cmidrule(lr){4-5}
    \cmidrule(lr){7-8}
    \cmidrule(lr){9-10}
    & Dx+Px & Rx
    & Dx+Px & Rx
    & Dx+Px 
    & Dx+Px & Rx
    & Dx+Px & Rx\\
    \midrule
    w/ AP, RC
    & 24.1 (0.8) & \textbf{44.6 (1.3)}
    & 39.6 (0.8) & 37.5 (1.9)
    & 84.6 (0.2)        
    & 20.8 (0.2) & 53.4 (0.1)   
    & \textbf{51.4 (0.6)}   & 56.0 (1.1)     
    \\
    w/ AP
    & 22.9 (1.7) & 43.9 (1.5)
    & 37.1 (1.2) & \textbf{41.9 (1.4)}
    & 84.4 (0.2)       
    & \textbf{20.9 (0.2)} & \textbf{54.7 (0.2)}   
    & 46.0 (0.3)   & \textbf{87.7 (0.1)}     
    \\
    w/ RC
    & 22.2 (1.0) & 39.4 (1.8)
    & \textbf{40.8 (2.0)} & 40.8 (4.0)
    & \textbf{84.7 (0.2)}       
    & 20.6 (0.3) & 53.8 (0.3)   
    & 14.0 (1.3)   & 13.6 (2.9)     
    \\
    -
    & \textbf{24.8 (0.6)} & 39.9 (1.4)
    & 38.6 (1.4) & 40.1 (1.5)
    & 83.9 (0.4)       
    & 20.7 (0.1) & 54.0 (0.3)   
    & 13.0 (0.4)   & 28.0 (2.9)     
    \\
    \bottomrule
    \end{tabular}
    }

    \label{tab:loss}
\end{table*} 

\subsection{Ablation Studies}
\label{ssec:abaltion}

\subsubsection{Effect of Alternative Design Choices}

We describe four variants of \modelvts and report their downstream task performances in Table~\ref{tab:ablation_model}.

\noindent \textbf{+\;init} \enskip is another way of infusing textual descriptions into the graph representation. 
\textbf{+\;init} replaces the graph embeddings with the mean of the textual description token embeddings.
In general, the summary network shows better performance, but \textbf{+\;init} outperforms on cross-modal retrieval. 
Unlike \model, \textbf{+\;init} has the advantage to capture the bag-of-(sub)words similarity between two modalities because \textbf{+\;init} directly injects the bag-of-words as input and keeps it along with a residual connection. Despite the importance of the bag-of-words similarity in the information retrieval, our model offers a flexible summary of textual descriptions in a graph, which leads to a performance improvement on other tasks.

\noindent \textbf{-\;structure} \enskip is used to test the importance of the structural information of EHR graph.
We destroy the structural information by concatenating all textual description $D$ into a single passage to represent the EHR graph.
As there is no structure we use Transformer instead of GAT to encode the passage.
For both Dx+Px and Rx, the structural information of a graph plays a crucial role for all tasks except note generation for Rx. 
This can be due to the Rx section of the discharge summary being more or less a permutation of textual descriptions $D$, making the note generation a rather straightforward problem for \textbf{-\;structure}.
However, \textbf{-\;structure} loses merit when the task is hard to solve with only textual descriptions $D$, such as generation on Dx+Px.

\noindent \model shows superior performance on three tasks and takes second place on the retrieval and generation. Given \model's stable performance in all tasks, we believe \model has a larger potential for generalizability than other baselines.

\begin{figure*}[t]
    \floatconts{fig:qual_sample}
    {\caption{
    Examples of the cross-modal attention distribution where the query is a graph node and the key is a text token. The left and right sides of each figure are text tokens and graph nodes, respectively.
    Different colors represent different attention heads. Best viewed in color.}}
    {\includegraphics[width=2.0\columnwidth]{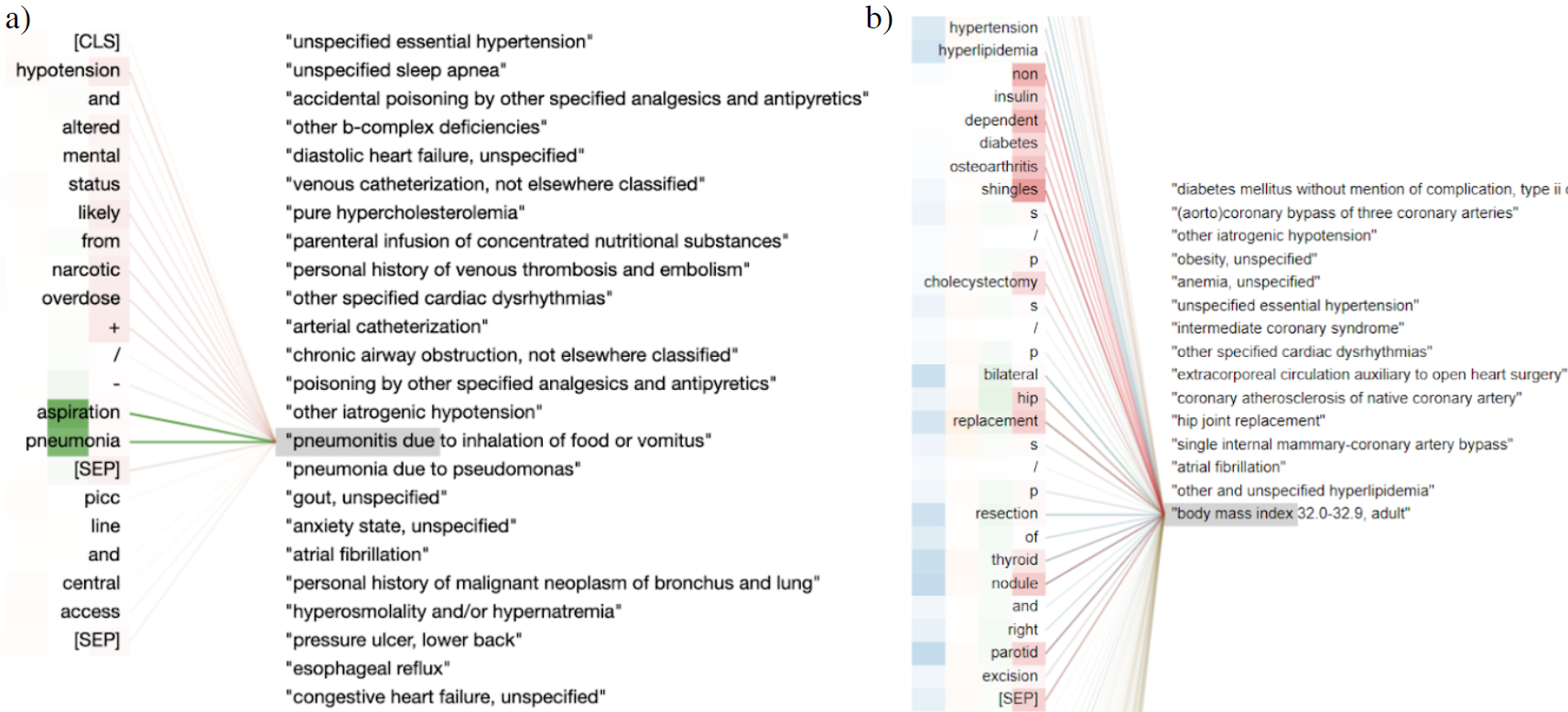}}
\end{figure*}

\subsubsection{Effect of model components}

Our design choices were motivated by a volume of recent multi-modal frameworks from other domains such as vision-language or code-text representation learning~\citep{tan2019lxmert, guo2020graphcodebert}.
To empirically justify our multi-modal framework, we evaluate the effect of two model components, namely the structure-aware attention mask and the text encoder initialization with BERT parameters. 
Here, we employ \model as a core model, which shows the impressive performance among the variants of our multi-modal framework. 
For \textbf{-\;\;GAT}, we replace the structure-aware attention mask $M_{adj}$ with the padding mask $M_{pad}$ of the graph encoder.
For \textbf{-\;\;LM}, we randomly initialize the text encoder instead of using the pre-trained weights from BERT. 
We report the downstream task performance in Table~\ref{tab:ablation_components}. Note that the performance of the error detection on Rx is saturated for all models, so we exclude the results from Table~\ref{tab:ablation_components}.

In all downstream tasks, the performance significantly drops when we do not initialize \model with BERT, so the pre-trained language model plays a crucial role in understanding the discharge summary. In the case of \textbf{-\;\;GAT}, it shows a comparable performance for mortality prediction on Rx, error detection on Dx+Px, and note generation while other task performances significantly decrease. \model always experiences the performance drop when we remove one of the components from the model, indicating the effectiveness of the proposed model architecture.

\subsubsection{Effect of pre-training tasks}
We establish masked language modeling (MLM) for the text part and masked literal prediction (MLP) for the graph part as major pre-training tasks. Therefore, we study the effect of the other two pre-training tasks, namely cross-modal alignment prediction (AP) and relation classification (RC), on the downstream tasks as shown in Table \ref{tab:loss}. 
For this experiment, we use \model as a base model.
Two additional pre-training objectives consistently show a positive impact on most datasets and downstream tasks except for mortality prediction on Rx, but their effect varies across datasets and tasks.
In the case of the Rx dataset, \model trained with AP consistently shows a comparable and outperforming performance on every downstream task.
For the Dx+Px dataset, training \model with AP and RC is a reasonable choice due to its stable performance over downstream tasks.
Furthermore, it can be seen that AP is essential for the cross-modal alignment task.

\subsection{Attention Visualization}
To indirectly evaluate the model's understanding of the multi-modal input, we can observe how one modality refers to another in the cross-attention layer.
We observe the three types of connection between two modalities: 1) same surface form, 2) different surface forms, but represent the same medical concept, and 3) different medical concepts, but capture a biomedical correlation between concepts.
We visualize the attention weights in Figure~\ref{fig:qual_sample}. In Figure~\ref{fig:qual_sample} a), the literal node \textit{pneumonitis due to inhalation of food or vomitus} strongly focuses on \textit{aspiration pneumonia} in text. This implies that cross-attention helps to align the same clinical concepts even they have a different surface form. 
In Figure~\ref{fig:qual_sample} b), literal node \textit{body mass index 32.0-32.9, adult} not only refers to its synonym but also other clinically meaningful texts such as \textit{hyperlipidemia}, \textit{diabetes}, and \textit{osteoarthritis}, which are closely related to the lifestyle disease (\textit{i.e.}, adult diseases).
Other examples and detailed explanations are in Appendix~\ref{supp:att_viz}.

\section{Conclusion}
\label{sec:conclusion}
This paper presents \modelvts, which learns joint representations of the structured and textual data of EHR. 
To the best of our knowledge, this is the first pre-trained model that treats both modalities equally compared to previous approaches that handle one modality as an auxiliary information source of another modality.
Focusing on the EHR specific nature, we proposed four pre-training tasks and a novel summary network which brought improved performance for the five clinical downstream tasks.
Further qualitative analysis and ablation study confirmed the efficacy of the proposed methods. One important future direction is to extend our model to entire EHR data (not just discharge summaries), which will lead to a true understanding of EHR, although extensive preprocessing and hardware resource requirements must be addressed at the same time.


\section*{Institutional Review Board (IRB)}
This research does not require IRB approval.

\acks{
This work was supported by Institute of Information \& Communications Technology Planning \& Evaluation (IITP) grant (No.2019-0-00075, Artificial Intelligence Graduate School Program(KAIST)), National Research Foundation of Korea (NRF) grant (NRF-2020H1D3A2A03100945), and the Korea Health Industry Development Institute (KHIDI) grant (No.HI21C1138), funded by the Korea government (MSIT, MOHW).
}

\bibliography{jmlr-sample}

\clearpage
\appendix
\section{Further discussion}
\label{supp:ablation}

\subsection{Abstract nodes \& literal nodes}
\label{supp:absLiteralNodes}
Unlike literal nodes whose vocabulary size is typically bounded (there is a finite number of unique medical codes), the vocabulary size of abstract nodes is, in theory, unbounded since new visits or new medication orders occur every day.
Also, abstract nodes do not mean anything in themselves but are simply an aggregation of information from their children nodes.
Therefore, to reduce the node space complexity and encourage efficient representation learning of the EHR graph, we replace each abstract node with a placeholder node.
Specifically, admission, diagnosis, procedure, and prescription nodes are replaced with [ADM], [DX], [PX], and [RX], respectively.

\begin{figure*}[t]
\floatconts
    {fig:table2kg_conversion}
    {\caption{Following~\citep{park2020knowledge}, five tables in MIMIC-III (\textit{i.e.} diagnoses, d\_icd\_diagnoses, procedures, d\_icd\_procedures, prescriptions) are converted to a knowledge graph. 
    Rectangles and ellipses in the graph indicate entities and literal values, respectively. We omit relations from the figure for simplicity.}}
    {\includegraphics[width=2.0\columnwidth]{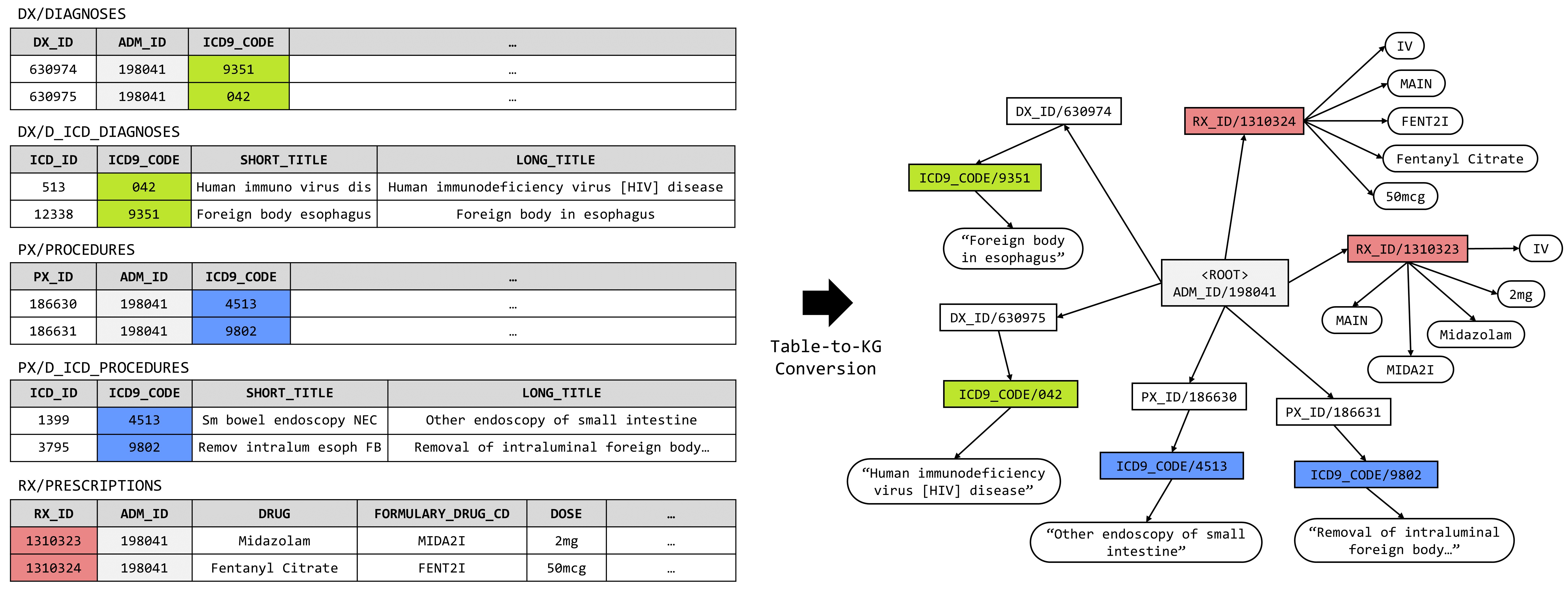}}
\end{figure*}

\section{Details of EHR Graph}
\label{supp:data_proc}
\subsection{Table to Knowledge Graph Conversion}
Knowledge Graphs (KG) are a set of triples that describe the relationship between two entities or between an entity and a literal value \citep{chakraborty2019introduction}. 
To extract triples from the tables of MIMIC-III,~\citep{park2020knowledge} defined an algorithm to map each column to either an entity, a literal, or a relation. 

By following the approach~\citep{park2020knowledge}, we first choose five tables from MIMIC-III: diagnoses, procedures, prescriptions, diagnosis descriptions, and procedure descriptions.
We then merge the procedures and their descriptions table into a single procedure table as well as a diagnoses table.
The primary or foreign key columns of a table are defined as entities, and non-key columns (i.e., property columns) are defined as literals. For example, in the prescriptions table of Figure~\ref{fig:table2kg_conversion}, the values of ADM\_ID are defined as entities that can have child nodes. The values of DRUG, FORMULARY\_DRUG\_CD, and DOSE are defined as literals that cannot have child nodes. 
The column names of the non-key columns become relations, e.g. /diagnoses\_long\_title links between ICD9\_CODE/4513 and “Other endoscopy of small intestine”. 
Lastly, the relation between a primary key and a foreign key is derived from the table name (e.g., /procedures links between ADM\_ID/198041 and PX\_ID/186630). 

\subsection{Relations in EHR graph}
For Dx+Px, relation types are /diagnoses\_icd9\_code, /diagnoses, /procedures, /procedures\_icd9\_code, /diagnoses\_long\_title, /procedures\_long\_title.
\\ 
For Rx, relation types are /icustay\_id, /prescriptions, /drug\_type, /drug, /route, /formulary\_drug\_cd, /drug\_dose.

\section{Experiment details}
\label{supp:exp_setup}

\subsection{Pretrain}
\label{supp:PreSetting}
\paragraph{Data preparation}
For the graph modality, we convert five MIMIC-III tables into the EHR graph as in~\ref{supp:data_proc}.
For the text modality, we extract clinical notes from noteevents table in MIMIC-III. Among various kinds of notes, our target is the discharge summary, so we divide the text belonging to the discharge summary based on section headers and extract each section separately.
The section headers corresponding to diagnoses, procedures, and prescriptions are discharge diagnosis, major surgical or invasive procedure, and discharge medication, respectively. 
The complete dataset is constructed by coupling a subgraph and text which belongs to the same admission id.
We drop the admission that has less than five words in the text modality to guarantee the existence of text information.
For Rx, we filter the subgraph which has more than 768 nodes including the admission node.
A limitation of existing Transformer models and their derivatives is that the full self-attention mechanism has computational and memory requirements that are quadratic with the input sequence length. To fit our model into RTX 3090 with a reasonable batch size (at least 4), the maximum possible sequence length was near 768. This design choice truncates 22\% of the samples in the dataset. We tried to increase the sequence length from 768 to 1024, and this change only restores the 5\% of samples while the model does not fit in GPU.

As shown in Section~\ref{sec:datasets} and Table~\ref{tab:data_stat}, we merge diagnosis (Dx), procedures (Px) into one dataset (Dx+Px) without any modifications.

\paragraph{Implementation details}
We construct the graph vocabulary with unique abstract, literal nodes and three special tokens for a graph modality ([SUM], [MASK]$_G$, [PAD]$_G$). We use the same vocabulary which BERT used for the text. For the model's input, the graph nodes are first listed in breadth first search (BFS) order followed by tokenized text subwords. We pad each modality inputs to the maximum length and add special tokens.

The two essential pre-training tasks, masked language modeling (MLM) and masked literal prediction (MLP), follow the standard masking strategy in BERT: randomly masking out 15\% of the tokens in the input, and pass the masked sequence through the model and predict the masked tokens. In MLP, only literal nodes are treated as masking targets, and both MLM and MLP use the cross entropy loss over the masked tokens as an objective function.
For relation classification (RC), we provide the position of two nodes and the type of relation between them as an input, and the outputs of two nodes are concatenated and pass through the multi-class classifier.
For cross-modal alignment prediction (AP), we follow the same setting in Section~\ref{ssec:pretrain}.
Note that RC and AP also use the cross entropy loss to optimize the model during pre-training.

\paragraph{Model hyperparameter settings} We set the hidden dimension of \modelvts and all baselines to 128. The number of blocks in each single modality encoder is 2, and 4 for the cross-modal encoder. The number of attention heads is always 4. 

\subsection{Cross-modal retrieval}
\label{supp:ReSetting}
\paragraph{Data preparation} We use the same data and splits in Section~\ref{supp:PreSetting}. We sample the negative pair inside the batch where the graph or text belongs to another admission.
\paragraph{Training \& Evaluation}
For fine-tuning, we concatenate [SUM], [CLS] together and feed them into two layers of feed-forward network with a non-linear activation function (\textit{tanh}) between those layers. The classifier predicts the cross-modal alignment score of the graph-text pair by taking softmax on the output. 

During the inference, we consequently sample the graph (text) from the test set, and measure the cross-modal alignment score with the entire text (graph) in the test set. We calculate the rank based on cross-modal alignment score where a higher cross-modal alignment score gets a higher position in rank.

\begin{table*}[ht]
\caption{Hyperparameters for each task.}
\resizebox{2.0\columnwidth}{!}{%
  \begin{tabular}{cccccc}
    \toprule
    Hyperparameters
    & Pre-training
    & Cross-modal retrieval
    & Temporal prediction
    & Error detection 
    & Clinical note generation
    \\
    \midrule
    \multirow[m]{1}{*}{Learning rate}
    & $1\times10^{-4}$
    & $1\times10^{-5}$
    & $1\times10^{-4}$
    & $1\times10^{-5}$
    & $3\times10^{-5}$ 
    \\
    \multirow[m]{1}{*}{Training epochs}
    & 40 
    & 20 
    & 20 
    & 20 
    & 30 
    \\
    \multirow[m]{1}{*}{Number of negative sample}
    & 1 
    & 1 
    & 1 
    & - 
    & - 
    \\
    \multirow[m]{1}{*}{MLM probability}
    & 0.15 
    & - 
    & - 
    & - 
    & 0.15 
    \\
    \multirow[m]{1}{*}{Corruption probability}
    & - 
    & - 
    & - 
    & 0.25 
    & - 
    \\
    \multirow[m]{1}{*}{$p$ (Top-$p$ sampling)}
    & - 
    & - 
    & - 
    & - 
    & 0.9 
    \\
    \multirow[m]{1}{*}{Dropout probability}
    & \multicolumn{5}{c}{0.1} 
    \\
    \multirow[m]{1}{*}{Batch size}
    & \multicolumn{5}{c}{16 (Dx+Px) / 8 (Rx)}
    \\
    \multirow[m]{1}{*}{Random seed}
    &\multicolumn{5}{c}{1, 12, 123, 1234, 42}
    \\
    \multirow[m]{1}{*}{Maximum length of text}
    &\multicolumn{5}{c}{512}
    \\
    \bottomrule
  \end{tabular}
  }
  \label{tab:hyperparams}
\end{table*}

\subsection{Temporal prediction}
\paragraph{Data preparation}
We select patient cohorts starting from the same data and splits in Section \ref{supp:PreSetting}.
For the readmission prediction task, each dataset is divided into two groups based on whether readmitted (1) or not (0) within 30 days after discharge.
Similarly, for the mortality prediction task, the target label depends on whether a patient is either dead within 30 days (1) or not (0).
For Dx+Px, the final cohorts for hospital readmission and mortality prediction contain 6,948 and 26,577 training samples with 2,072 and 1,439 positive labels, respectively.
For Rx, for the same downstream tasks, the final cohorts contain 5,808 and 23,412 training samples with 1,597 positive labels and 1,108 positive labels, respectively.

\paragraph{Training \& Evaluation}
For training, we first pass the inputs to the model and then concatenate the output vectors which correspond to [SUM], [CLS] together and feed them into two layers of feed-forward network with a non-linear activation function (\textit{tanh}) between those layers. For the temporal prediction task, we use binary classifier outputs via sigmoid operations. We optimize the model by minimizing binary cross entropy loss over predicted and ground truth labels.

\subsection{Error detection}
\label{supp:EDSetting}
\paragraph{Data preparation} 
We randomly replace the literals in the graph with a probability of 0.25.
For Rx graphs, we alter five types of literal nodes (\textit{i.e.} drug name, drug code, route, formulary drug cd, and dosage).
For Dx+Px graphs, we alter two types of literal nodes (\textit{i.e.} diagnosis name, procedure name). 
We choose the alternative that belongs to the same type as the original node. 

\paragraph{Training \& Evaluation}
We add a one-versus-all (OVA) binary classification head on the top of \modelvts. The OVA classifier receives a contextualized embedding of each node as input and predicts whether each node in the EHR graph is altered (1) or not (0). To be specific, the OVA classifier consists of a simple linear layer and optimizes the model by using binary cross-entropy loss.
For evaluation metric, we use F1 score to report the model performance.

\subsection{Clinical note generation}
\label{supp:GenSetting}
\paragraph{Data preparation}
We use the same data and splits in Section~\ref{supp:PreSetting}. To directly leverage source database (\textit{i.e.} MIMIC-III) information for the generation task, we do not use specialized meta-tags or pre-processing techniques.

\paragraph{Training \& Evaluation}
Following the fine-tuning strategy in UniLM~\citep{NEURIPS2019_c20bb2d9}, we fine-tune the model on masked language modeling task, but the attention mask is specialized for a sequence-to-sequence (S2S) task. This S2S attention mask adds the causality constraint to the text modality while the graph nodes can attend any other tokens (include itself) in a graph. Note that it's impossible to maintain uni-directionality in a text modality with a cross attention mechanism because the bidirectional information of the text flows through the previous cross-modal attention layers. Therefore we cut off one of the cross-modal attention (\textit{i.e.} $\text{CrossAttn}_{G \rightarrow T}$), and use a causal attention mask for the text modality. We also masked the [SEP] token with a masking probability of 0.5 to encourage the model to emit [SEP] token which terminates the generation process.

During the evaluation, we first pack the whole graph inputs and [CLS] into a sequence. We append [MASK] token to the last of a sequence and predict the masked token. We run this procedure in an auto-regressive and greedy manner to decode a sequence until the [SEP] token comes out. For the Dx+Px dataset, we keep decoding until the second [SEP] token is emitted. Also, we change section embeddings after the model emits the first [SEP] token. 
Since we choose the optimal decoding strategy over all models for each dataset, we adopt top-$p$ sampling strategy for Rx.

\section{Cross attention visualization}
\label{supp:att_viz}
In this section, we visualize the additional samples which belong to one of three types of connection between two modalities:
1) same surface form (in Figure~\ref{supp:attn_qual_1});
2) different surface forms, but represent the same medical concept (in Figure~\ref{supp:attn_qual_2},\ref{supp:attn_qual_3});
3) different medical concepts, but capture a biomedical correlation between concepts (in Figure~\ref{supp:attn_qual_4}).

\section{Important hyperparameters}
We specify hyperparameter settings including pre-training and all downstream tasks in Table~\ref{tab:hyperparams}.

\section{Experiment environments}
We train our models on NVIDIA GeForce RTX 3090 and RTX A6000. 
Also, We use CUDA 11.1, and PyTorch 1.9.0. for experiments.

\begin{figure*}[h]
\floatconts
    {supp:attn_qual_1}
    {\caption{
    \textbf{Same surface form};
    The node ``\textit{urinary tract infection, site not specified}'' (graph) highlights \{``\textit{urinary}'', ``\textit{tract}'',  ``\textit{infection}''\} (text) in the cross attention map.
    Although the \modelvts does not use information about the surface form of words, it can be seen that attention weights are strongly applied to words with the same meaning.
    }}
    {\includegraphics[width=2.0\columnwidth]{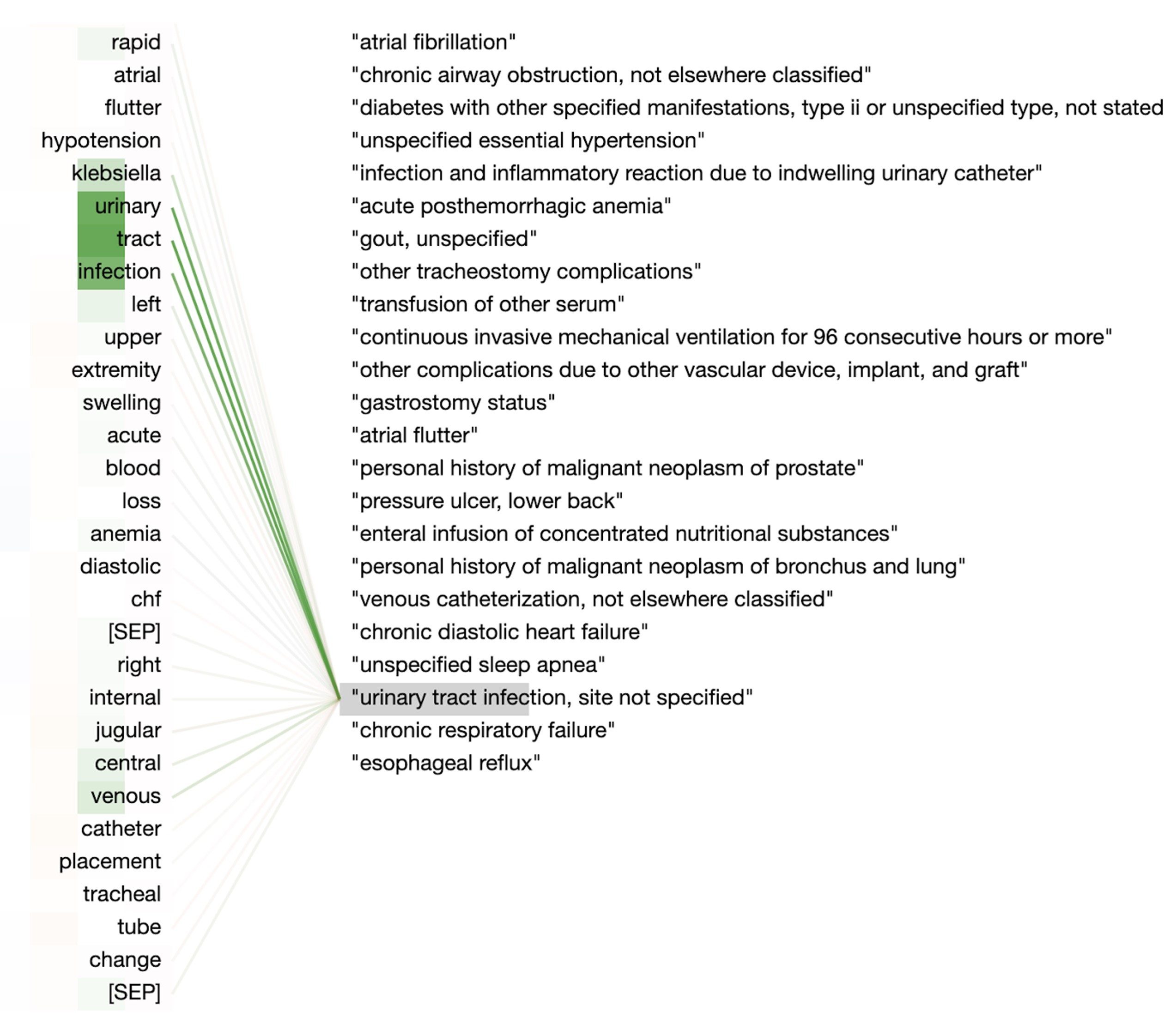}}
\end{figure*}

\begin{figure*}[h]
\floatconts
    {supp:attn_qual_2}
    {\caption{
    \textbf{Different surface forms, but represent the same medical concept};
    The token ``\textit{uti}'' (text) highlights 
    ``\textit{urinary tract infection, site not specified}'' (graph) in the cross attention map.
    ``\textit{uti}'' is an abbreviation of ``\textit{urinary tract infection}''. They have different surface forms but have the same meaning.
    }}
    {\includegraphics[width=2.0\columnwidth]{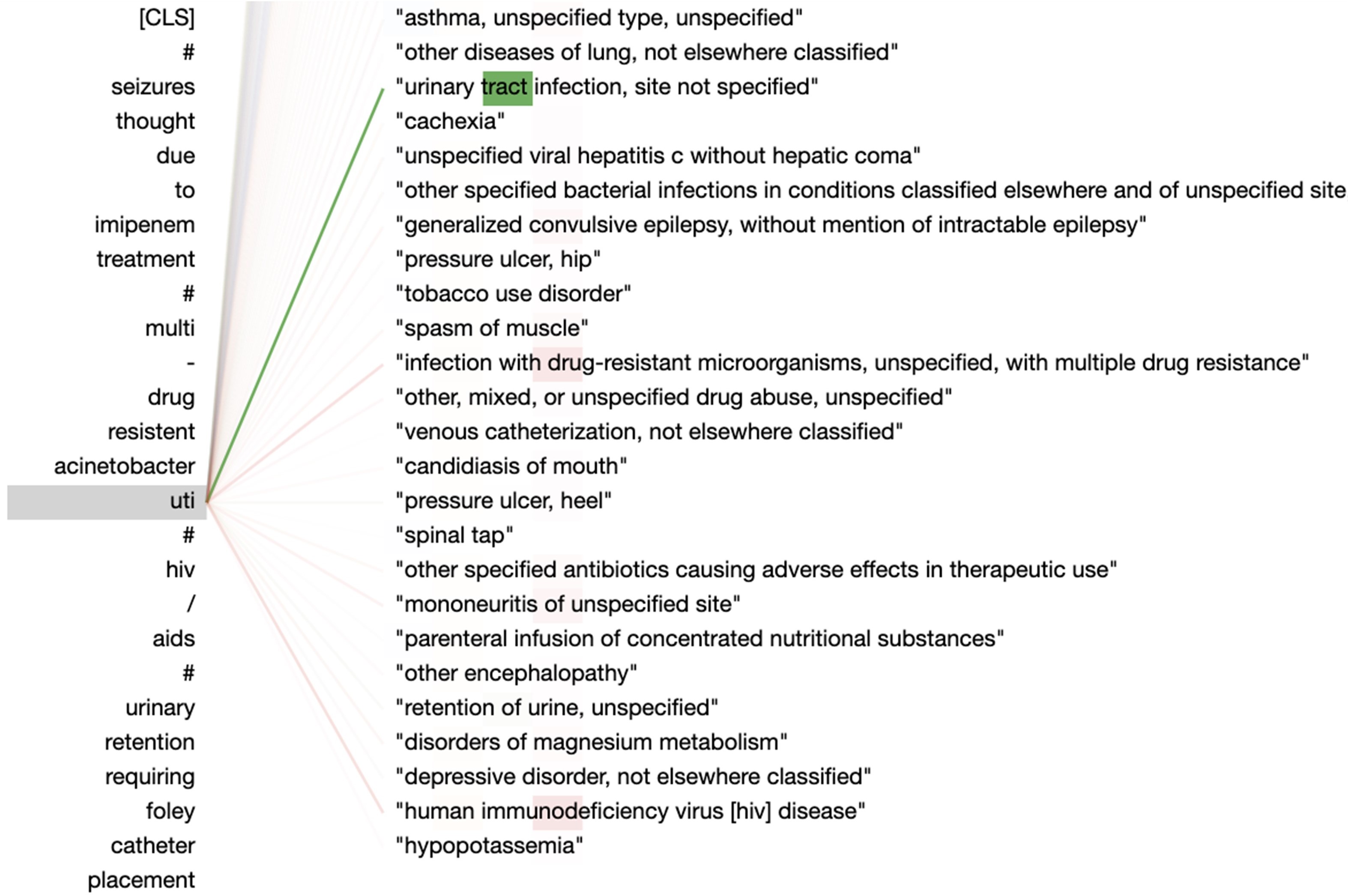}}
\end{figure*}

\begin{figure*}[h]
\floatconts
    {supp:attn_qual_3}
    {\caption{
    \textbf{Different surface forms, but represent the same medical concept};
    The node ``\textit{acute kidney failure, unspecified}'' (graph) highlights
    \{``\textit{acute}'', ``\textit{renal}'', ``\textit{failure}''\} (text) in the cross attention map.
    ``\textit{renal}'' differs from ``\textit{kidney}'' in the appearance, but they have the same meaning.}}
    {\includegraphics[width=2.0\columnwidth]{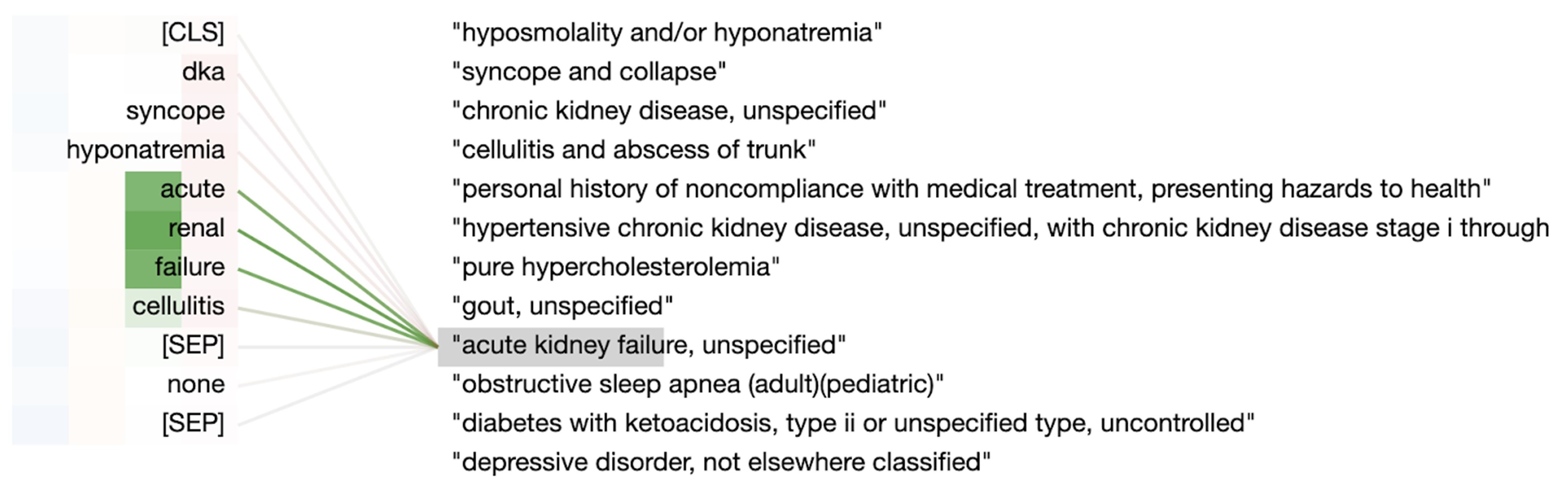}}
\end{figure*}

\begin{figure*}[h]
\floatconts
    {supp:attn_qual_4}
    {\caption{
    \textbf{Different medical concepts, but capture a biomedical correlation between concepts}; The token
    ``\textit{mv}'' (text)
    highlights
    ``\textit{mitral valve disorders}'' (graph) in the cross attention map.
    ``\textit{mv}'' is an abbreviation for ``\textit{mitral valve}''. They have different surface forms but have the same meaning. Furthermore, ``\textit{mv}'' also strongly attends on  ``\textit{open heart valvuloplasty of mitral valve without replacement}'' and ``\textit{extracorporeal circulation auxiliary to open heart surgery}'' nodes. They have different medical concepts, but it captures a biomedical correlation between concepts.}}
    {\includegraphics[width=2.0\columnwidth]{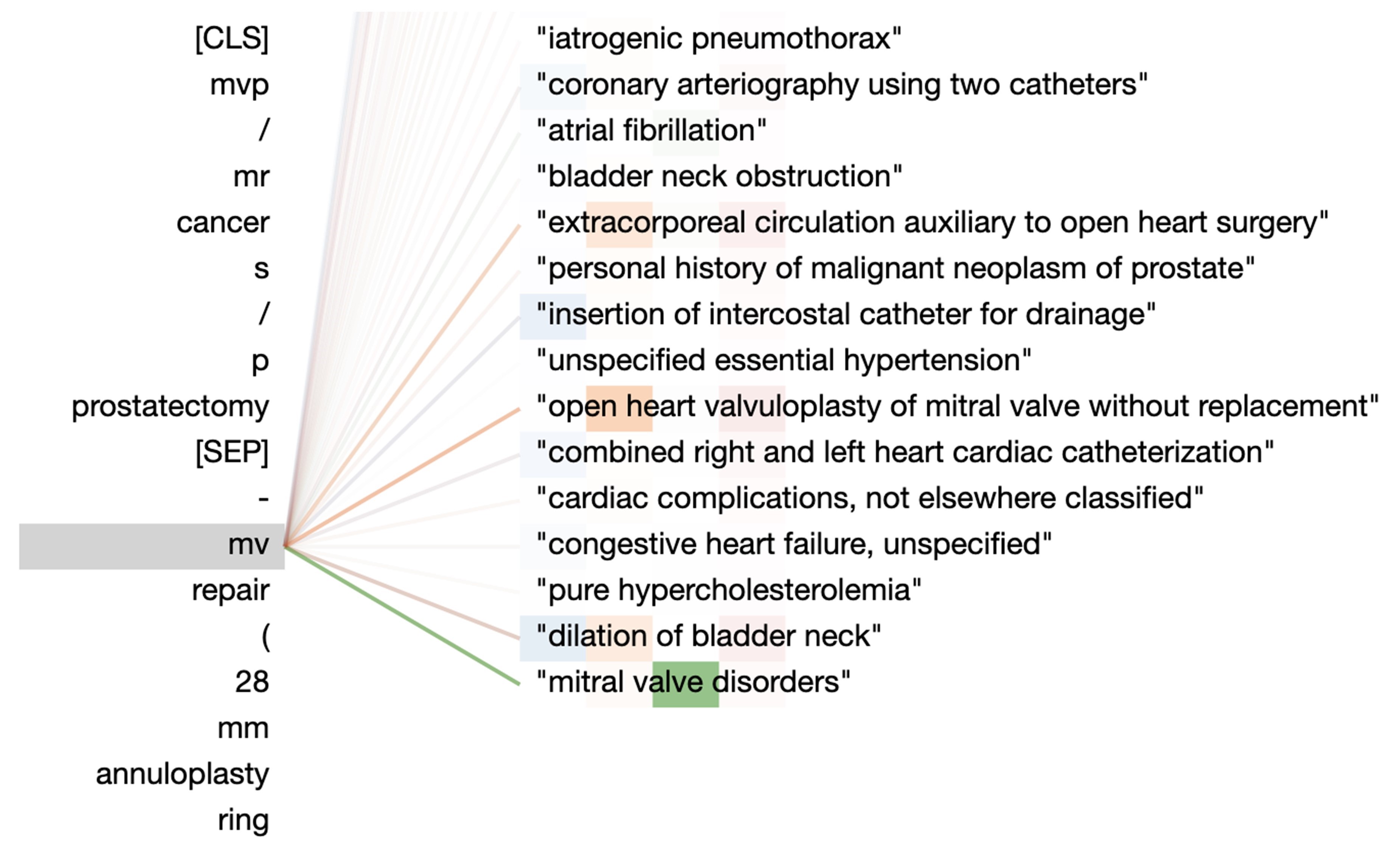}}
\end{figure*}



\end{document}